\documentclass{article}

\usepackage[preprint, nonatbib]{neurips_2024}

\usepackage[utf8]{inputenc} % allow utf-8 input
\usepackage[T1]{fontenc}    % use 8-bit T1 fonts
\usepackage{hyperref}       % hyperlinks
\usepackage{url}            % simple URL typesetting
\usepackage{booktabs}       % professional-quality tables
\usepackage{amsfonts}       % blackboard math symbols
\usepackage{float} 
\usepackage{nicefrac}       % compact symbols for 1/2, etc.
\usepackage{microtype}      % microtypography
\usepackage{subcaption}
\usepackage{xcolor}         % colors
\usepackage{graphicx}
\usepackage{array}
\usepackage{amsmath}
\usepackage{pdfpages}

\title{Learning Physical Simulation with \\ Message Passing Transformer}

\author{
  Zeyi Xu\thanks{Email: \texttt{xzyblxa@shu.edu.cn}} \\
  Shanghai University \\
  \and
  Yifei Li$^\dagger$ \\
  MIT CSAIL
}

\begin{document}

\maketitle
\begin{abstract}
    Machine learning methods for physical simulation have achieved significant success in recent years. We propose a new universal architecture based on Graph Neural Network, the Message Passing Transformer, which incorporates a Message Passing framework, employs an Encoder-Processor-Decoder structure, and applies Graph Fourier Loss as loss function for model optimization. To take advantage of the past message passing state information, we propose Hadamard-Product Attention to update the node attribute in the Processor, Hadamard-Product Attention is a variant of Dot-Product Attention that focuses on more fine-grained semantics and emphasizes on assigning attention weights over each feature dimension rather than each position in the sequence relative to others. We further introduce Graph Fourier Loss (GFL) to balance high-energy and low-energy components. To improve time performance, we precompute the graph's Laplacian eigenvectors before the training process. Our architecture achieves significant accuracy improvements in long-term rollouts for both Lagrangian and Eulerian dynamical systems over current methods.
\end{abstract}

\section{Introduction}
%===========================介绍GNN4Sim================================
In recent advancements in physical systems simulation, an increasing number of neural network-based methods are challenging traditional numerical solvers. These methods, notable for being several orders of magnitude faster and maintaining low error rates, have garnered substantial attention from researchers
\cite{NeuralODE}
\cite{FNO}
\cite{PINN}
\cite{GNS}
. Graph Neural Networks (GNNs) have garnered widespread interest and research attention due to their unique features, including node-wise independent updates and aggregation operations that closely resemble iterations in traditional simulations. Additionally, particle-based and mesh-based simulations can be easily converted into graphs through neighbor interactions \cite{GNS}  and topology structures \cite{MGN}. This growing body of research can be collectively termed as GNN for Simulation (GNN4Sim)
\cite{47094}
\cite{SanchezGonzalez2018GraphNA}
\cite{belbute2020combining}
\cite{GraphUNet}.

%===========================介绍EPD================================
Unlike the fields of Natural Language Processing (NLP) and Computer Vision (CV), physical systems are more complex and unstable, where every piece of information is crucial. This implies that issues like over-smoothing and over-squashing might be more pertinent in physical simulations. A common solution to mitigate these issues is mapping the original information to a high-dimensional latent space for processing. This approach has become a widely used architecture in most GNN4Sim applications
\cite{BSMS, MGN, MSMGN, GNS, C-GNS}
, known as the encoder-processor-decoder (EPD), which, due to its focus on nodes and edges rather than graph structure, has a broad range of applications.

%===========================介绍Transformer和改进================================
Since the Attention mechanism achieved significant success in the Natural Language Processing (NLP) domain
\cite{Transformer}
, many Transformer-based network architectures have been developed in computer vision
\cite{ViT, SwinT}
, and Social and Information Networks
\cite{GTN}
. To enhance the capabilities of EPD and to leverage the fixed number of message passing iterations, we replace the commonly used Multi-Layer Perceptron (MLP) with Hadamard-Product Attention. This variant of the Dot-Product Attention mechanism is tailored for inputs with a finite maximum sequence length, enabling more nuanced processing of the relative importance of features within the message passing framework of the processor.

%===========================介绍GFL================================
Incorporating theoretical methodologies from graph signal processing, we have innovatively applied the Graph Fourier Transform (GFT) 
\cite{6409473, hammond2011wavelets}
to the domain of physical simulations through our introduction of Graph Fourier Loss (GFL). This novel loss function optimizes model performance by leveraging the unique spectral properties of graphs. Rather than applying a Fourier Transform directly to the model, we employ preprocessing Laplacian eigenvector matrix within the loss function, thereby keeping the model's inference time nearly unchanged.
\section{Related Work}
%===========================介绍AI4Sim================================
\paragraph{Neural Approaches in Physical Systems Simulation}
In recent years, Several neural-based methods have achieve great research interest in physical systems simulation due to their high computational speed and low error rates. Among these, Convolutional Neural Networks (CNNs) 
\cite{Afshar2019PredictionOA, zhang2018application, _zbay_2021, tompson2017accelerating, CNNx, liu2023fast}
have been utilized to infer the dynamics of physical systems, demonstrating the capability of neural networks in accurately and efficiently simulating complex phenomena.
Physics-Informed Neural Networks (PINNs)
\cite{PINN, cuomo2022scientific, app10175917, wang2024pinn, wang2021understanding}
, which leverage implicit representations and physical condition constraints, can be trained without traditional datasets, illustrating a groundbreaking approach to model training that is particularly valuable in scenarios where empirical data is scarce or difficult to obtain.
Neural operators
% FNO, DeepONet, Neural operator
\cite{FNO, Lu2019LearningNO, kovachki2023neural, gupta2021multiwavelet}
offer a novel methodology for predicting the physical state at any given time step directly from initial conditions. This represents a significant shift from traditional simulation methods, enabling more efficient and flexible simulations across various scales and conditions.
Graph Neural Networks (GNNs)
% GNS
\cite{GNS, MGN, GraphCast, sanchez2018graph}
facilitate information exchange between nodes through message passing. This mechanism effectively captures the interactions within physical systems, allowing for the detailed simulation of complex dynamics.

%===========================介绍EPD架构改进================================
\paragraph{Advancements in GNN Architectures for Simulation}
Recent studies in GNN4Sim have introduced various improvements. Multiscale methods
\cite{BSMS, MSMGN, Liu2023FastFS, Barwey2023MultiscaleGN}
, benefiting from simplified latent graph structures, have significantly accelerated training and inference while maintaining quality. At its core, this involves enhancing the efficiency of message passing, which entails modifying the topological structure of the graph. This change aims to concurrently increase processing speed and enhance accuracy in the simulations. Similarly, FIGNet
% FIGNet
\cite{FIGNet}
by adding face-face edges, changes the graph structure to improve collision accuracy. Han et al.
% temporal attention
\cite{temporal_attention}
 through uniform sampling, simplifies the graph structure, applying scaled dot-product attention to the entire graph but requires multiple prior temporal steps information. TIE
 \cite{TIE}
streamlines interaction modeling in Message Passing Neural Networks, utilizing a modified attention mechanism to efficiently process particle dynamics without explicit edge representations. LAMP
% LAMP
\cite{LAMP}
uses reinforcement learning to adapt to the varying relative importance of the trade-off between error and computation at inference time. C-GNS
% C-GNS
\cite{C-GNS}
focus on model the constraints of the physical system.

%===========================介绍GSP================================
\paragraph{Bridging Graph Theory and Signal Processing}
Graph Signal Processing (GSP) extends traditional signal processing techniques to signals defined on graphs
\cite{von2007tutorial}
\cite{Chung1997}
, a paradigm shift that has unlocked new avenues in analyzing complex data structures. Bruna et al.
\cite{bruna2013spectral}
 introducing Spectral Networks for graph data learning,  establishing foundational techniques for GNNs.
% Wavelets on Graphs via Spectral Graph Theory
Sandryhaila and Moura
\cite{6409473}
and Hammond et al.
\cite{hammond2011wavelets}
introduced the concept of applying wavelet transforms on graphs, offering a powerful tool for signal analysis and processing on irregular domains. Further, the development of Graph Convolutional Networks (GCNs)
% GCNs, FastGCN
\cite{GCN, Chen2018FastGCNFL, defferrard2016convolutional}
, simplified the application of convolutional neural networks to graph data, enabling efficient learning of graph-structured data. These foundational studies emphasize the significance of spectral methods in understanding and leveraging the inherent structure of data represented as graphs.
\section{Problem Formulation and Preliminaries}
This section introduces the problem formulation and the necessary preliminary concepts. It begins with the representation of physical systems using graph structures and the optimization goal for a learnable simulator in Section \ref{sec:PF}. It then delves into the Graph Fourier Transform (GFT), which facilitates the analysis of graph signals in the spectral domain in Section \ref{sec:GFT}.

\subsection{Problem Formulation}
\label{sec:PF}
We consider graph \( G^t = (V^t, E^t) \) to represent a physical system with \( t \) taking discrete values \( t = 0, 1, \ldots \), where \( V \) denotes the set of nodes with node attributes \( v_i \) for each \( v_i \in V \), and \( E \) denotes the set of edges with edge attributes \( e_{ij} \) for each \( e_{ij} \in E \). The graph is characterized by a fixed number of nodes \( N \), and the mesh topology of the corresponding graph \( G \) does not vary over time. This implies a constant structure where the connectivity pattern among nodes is preserved across all time steps.
We also define a total of \( M \) Message Passing iterations, with \( k = 0, 1, \ldots, M \). During the \( k \)-th Message Passing iteration, the attributes of nodes and edges are denoted by \( v_{k,i} \) and \( e_{k,ij} \), respectively, for each \( v_i \in V \) and \( e_{ij} \in E \).

The learnable simulator \( f_\theta \), parameterized by \( \theta \), can be optimized towards training objective. The goal of the learnable simulator is to predict the next state of the system, \( G^{t+1} \), based on the previous prediction of graph \( G^t \) at time step \( t \), denoted by \( G^{t+1} = f_\theta(G^{t}) \), or \( G^0 \rightarrow G^1 \rightarrow \cdots \rightarrow G^t \). 

\subsection{Graph Fourier Transform}
\label{sec:GFT}
The Graph Fourier Transform (GFT) transforms signals on a graph from the spatial vertex domain to the spectral frequency domain. For a signal defined on the vertices of the graph, GFT leverages the eigenvectors of the graph's Laplacian matrix, projecting the signal onto the orthogonal basis formed by these eigenvectors. This projection allows us to analyze and process the signal in a domain where convolution and filtering can be performed algebraically.
\subsubsection{Mathematical Definitions}
The adjacency matrix of $G$ is denoted by $A$, where $A_{ij}$ represents the weight of the edge between vertices $i$ and $j$. The degree matrix $D$ is a diagonal matrix where $D_{ii} = \sum_j A_{ij}$. The Laplacian matrix of the graph is defined as $L = D - A$.

The eigenvalues and eigenvectors of $L$ are denoted by $\lambda_i$ and $u_i$, respectively, where $i = 1, 2, \ldots, N$, and $N$ is the number of vertices in the graph. 

Given a signal $x \in \mathbb{R}^N$ defined on the vertices of the graph, the GFT of $x$ is given by
\[
\hat{x} = U^T x
\]
where $U = [u_1, u_2, \ldots, u_N]$ is the matrix of eigenvectors of $L$, and $U^T$ is its transpose. The signal $x$ can be reconstructed from its GFT $\hat{x}$ using the inverse GFT, given by
\[
x = U \hat{x}
\]
\section{Method}
\label{sec:Method}

\begin{figure}[htbp]
\centering
\includegraphics[page=3, width=\textwidth]{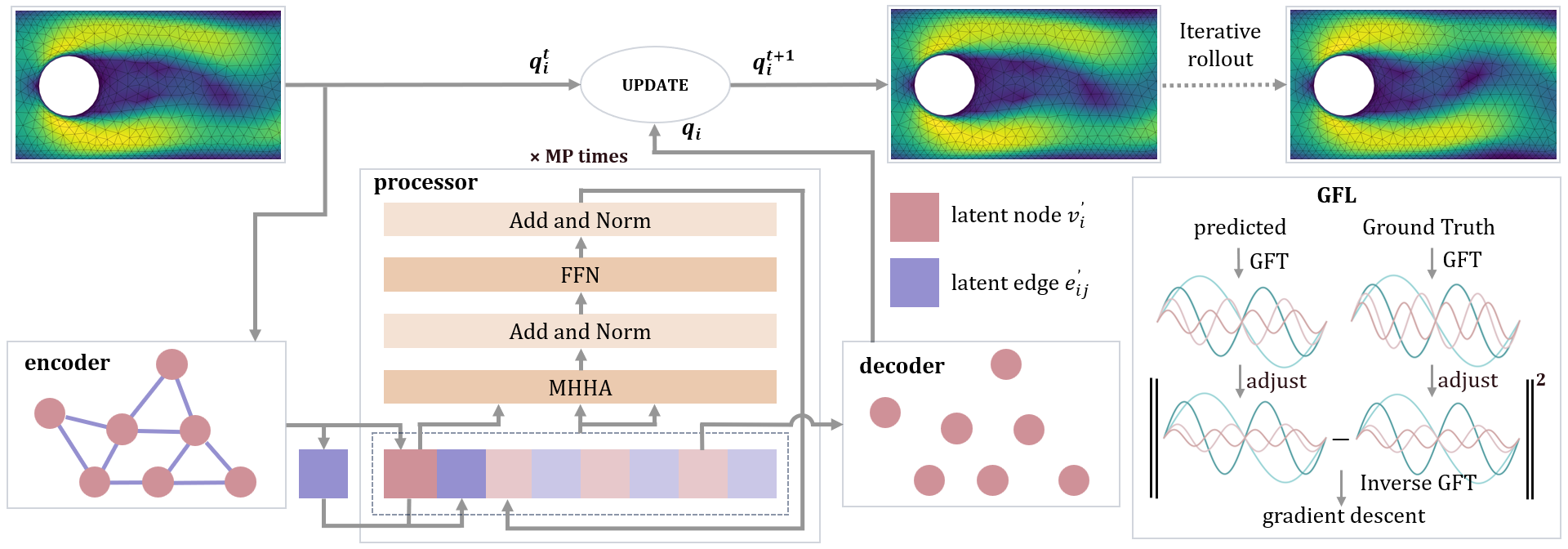}
\caption{Model Architecture of the Message Passing Transformer, visualizing the information processing procedure for the first of four Message Passing (\( k=0, M=4 \)) times. The encoder module transposes inputs into a latent space and the decoder predicts future states by extrapolating these encoded representations. The processor unit conducts numerous iterations, each treated as a regression problem, to refine node and edge attributes.}

\label{fig:model_architecture}
\end{figure}

In Section \ref{sec:MPT}, we present the overall architecture of the model, followed by a detailed description of the Hadamard-Product Attention and Graph Fourier Loss in Sections \ref{sec:HA} and \ref{sec:GFL}, respectively. The Hadamard-Product Attention is introduced to address aggregation bias and enable more fine-grained feature processing. The Graph Fourier Loss is introduced to balance the high-energy and low-energy components in the spectral domain, thereby enhancing the model's capacity to learn complex physical phenomena.

\subsection{Message Passing Transformer}
\label{sec:MPT}
The Message Passing Transformer architecture incorporates a Message Passing framework, employs an Encoder-Processor-Decoder structure, and utilizes Graph Fourier Loss for model optimization. Figure. \ref{fig:model_architecture} visualizes the computational process of the model. 
 
\paragraph{Encoder} The node and edge attributes are transformed into a latent space by \( f_1 \) and \( f_2 \), respectively. 
\[
v_{0,i} \leftarrow f_1(v_i), \quad e_{0,ij} \leftarrow f_2(e_{ij})
\]   
\paragraph{Processor}The edge features are updated by \( f_3 \), incorporating features from adjacent nodes. Node features are then updated by \( f_4 \), which aggregates information across multiple tokens using Hadamard-Product Attention. Each token represents node and aggregated edge features from a particular Message Passing iteration:
\[
e_{k+1,ij} \leftarrow f_3(e_{k,ij}, v_{k,i}, v_{k,j}), \quad v_{k+1,i} \leftarrow f_4\left( v_{k,i}, \bigoplus_{m=0}^{k} \left( v_{m,i}, \sum_{j} e_{m,ij} \right) \right)
\]
where \( \bigoplus \) denotes the sequential concatenation of tokens \( \left( v_{m,i}, \sum_{j} e_{m,ij} \right) \) for each \( m \) from 0 to \( k \), forming the input sequence for \( f_4 \).

\paragraph{Decoder}After \( M \) times Message Passing, the latent node features are mapped back to the original attribute space by \( f_5 \), culminating in the update of the graph state to the next time step.
\[
v_i' \leftarrow f_5(v_{M,i}), \quad G^{t+1} = \text{UPDATE}\left(G^t, v_i'\right)
\]
 
Here, \( f_1 \), \( f_2 \), \( f_3 \), and \( f_5 \) are all shallow MLPs. During training, finally, we compute the model loss using our Graph Fourier Loss and update the weights accordingly.

\subsection{Hadamard-Product Attention}
\label{sec:HA}
The Message Passing Transformer conceptualizes node updates as an autoregressive problem, utilizing past message-passing attributes for keys (K) and values (V), with the current state as the query (Q). 

\paragraph{Scaled Hadamard-Product Attention}
To address effectively sidestepping the aggregation bias introduced by the summation operations typical of matrix multiplication, we introduce the Scaled Hadamard-Product Attention (HPA) mechanism strategically utilizes element-wise multiplication followed by a linear transformation to finalize the attention computation.

Our Attention mechanism is tailored for inputs with a finite maximum sequence length, enabling more nuanced processing of the relative importance of features within the message passing framework of the processor. Assume query matrix \( {Q} \in \mathbb{R}^{b \times d_k} \), key matrix \( {K} \in \mathbb{R}^{b \times s \times d_k} \), and value matrix \( {V} \in \mathbb{R}^{b \times s \times d_k} \), where \( b \) represents the batch size, \( s \) denotes the sequence length, and \( d_k \) signifies the feature dimension. Particularly, \( Q \) represents the feature vector of an individual node. \( K \) and \( V \) constitute interlaced sequences where node attributes and aggregated edge attributes alternate. By performing broadcasting and the Hadamard product between the query matrix \( {Q} \) and the key matrix \( {K} \), we obtain the score matrix \( {S} \).

The Hadamard product is executed to compute the score matrix \( S \). Thus, for each \( i, j, k \), we have:
\[
{S}_{i,j,k} = \frac{{Q}_{i,j,k} \circ {K}_{i,j,k}}{\sqrt{d_k}}
\]
where \( Q \) is expanded from a shape of \( b \times d_k \) to \( b \times s \times d_k \), such that for all \( i \in \{1, \dots, b\} \), \( j \in \{1, \dots, s\} \), and \( k \in \{1, \dots, d_k\} \), the expanded \( {Q}_{i,j,k} = {Q}_{i,k} \). \( {S} \in \mathbb{R}^{b \times s \times d_k} \) represents the result of element-wise multiplication between each query and key. Notice that if calculating the sum of the feature dimensions in \( {S} \), the score matrix is identical to that of the Dot-Product Attention. This approach circumvents the summation aggregation inherent in matrix multiplication, thereby providing a more fine-grained representation.

Our attention mechanism that diverges from the traditional softmax application along the sequence dimension in \( S \), opting instead for execution along the feature dimension. 
This attention methodological shift implies that, contrary to the Scaled Dot-Product Attention, the computation of attention weights places emphasis on the allocation of weights across each feature dimension rather than assigning a weight to each position in the sequence relative to others. 
Consequently, the contribution of each dimension to the Value vector's computation is determined by its relative importance across the dimension, not by its position within the sequence.
\[
{attn}_{i,j,k} = \frac{\exp({S}_{i,j,k})}{\sum_{k'=1}^{d_k} \exp({S}_{i,j,k'})}
\]
Subsequently, a Hadamard product between the attention weights \( attn \) and the value matrix \( V \) is performed, resulting in a weighted value matrix \( w \in \mathbb{R}^{b \times s \times d_k} \).
\[
{w}_{i,j,k} = {attn}_{i,j,k} \circ {V}_{i,j,k}
\]
To finalize the transformation and align the output dimensions with the input, this weighted value matrix undergoes a linear transformation to reshape \( w \) from \( b \times s \times d_k \) back to a dimension of \( b \times d_k \).

\paragraph{Multi-Head Hadamard-Product Attention}
Incorporating the principles of Scaled Hadamard-Product Attention, we introduce the Multi-Head Hadamard-Product Attention (MHHA) as an adaptation of the conventional Multi-Head Attention mechanism. MHHA modifies the traditional approach by applying the Hadamard-Product Attention mechanism within each head, thereby facilitating a more nuanced processing of information across multiple representation subspaces.
\[
{Q}_i = {Q}W_i^Q, \; {K}_i = {K}W_i^K, \; {V}_i = {V}W_i^V
\]
where \(W_i^Q \in \mathbb{R}^{d_{model} \times d_k}\), \(W_i^K \in \mathbb{R}^{d_{model} \times d_k}\), and \(W_i^V \in \mathbb{R}^{d_{model} \times d_k}\) are the weight matrices for the \(i\)-th head. Scaled Hadamard-Product Attention is then applied to each head's transformed inputs, capturing detailed interactions within that subspace and finally concatenated and passed through a linear transformation \(W^O \in \mathbb{R}^{(d_k \times s) \times d_k} \):
\[
{head}_i = \text{HPA}({Q}_i, {K}_i, {V}_i)
\]

\[
\text{MHHA}(Q, K, V) = \text{Concat}({head}_1, ..., {head}_h)W^O
\]

\subsection{Graph Fourier Loss}
\label{sec:GFL}
Graph Fourier Loss (GFL) leverages the spectral properties of graphs to enhance model infering efficacy. By incorporating the graph's Laplacian eigenvectors in preprocessing and employing a unique loss function during training, GFL aims to mitigate computational burdens while improving model performance.

\paragraph{Preprocessing}
When the model does not alter the graph's topological structure, the inherent topological properties of the graph, such as the Laplacian matrix and its eigenvalues and eigenvectors, remain unaltered throughout the training process. To avoid the substantial increase in computation time caused by calculating eigendecompositions during training, we preprocess the training set before commencing model training. For each trajectory, the graph's Laplacian matrix is calculated and subsequently decomposed into eigenvectors \( U \) , which are stored and utilized during training.
% This approach eliminates the need for eigendecomposition of high-dimensional matrices during training, thereby maintaining consistent training times for the model.

\paragraph{Compute Graph Fourier Loss during training}
To circumvent the significant computational overhead of calculating eigenvectors during inference, we propose the Graph Fourier Loss (GFL) as the loss function. This strategy ensures the inference speed of the model remains unaffected. 

Initially, we perform GFT on both the model's output \(y^{\text{train}} \in \mathbb{R}^{N \times d_k}\) and the target output \(y \in \mathbb{R}^{N \times d_k}\), transforming the signals from the time domain to the frequency domain:
\[
\hat{y} = U^T y, \quad \hat{y}^{\text{train}} = U^T y^{\text{train}}
\]
Subsequently, we calculate the energy of each dimension of the transformed signals and sum them up to obtain the total energy for each signal across all nodes and dimensions:
\[
E = \sum_{k=1}^{d_k} |\hat{y}_{:,k}|^2, \quad E_{\text{train}} = \sum_{k=1}^{d_k} |\hat{y}^{\text{train}}_{:,k}|^2
\]
\(E\) and \(E_{\text{train}}\) represent the total energy of the target and model output signals in the frequency domain, respectively.

The energy \(E\) are then sorted, and using the hyperparameter segment rate \(s_r\), it is divided into high \(E_{\text{high}}\) and low \(E_{\text{low}}\) energy components. An adjustment factor \(\alpha\) is computed based on the mean energy of these partitions:
\[
\alpha = \sqrt{\frac{\text{mean}(E_{\text{low}})}{\text{mean}(E_{\text{high}}) + \epsilon}} \cdot \lambda
\]
The constant \(\epsilon\) is employed to prevent division by zero, while the regularization parameter \(\lambda\) controls the strength of the adjustment. When \(\alpha > 1\), high-energy regions are amplified, resulting in the model emphasizing high-energy components. Conversely, when \(\alpha < 1\), this emphasis is reduced. Both manual setting of the regularization parameter \(\lambda\) and incorporation it as a learnable parameter have been tested, found in Experiment \ref{par:ablation_studies}

Finally, we adjust the signals and compute the mean squared error (MSE) directly in the frequency domain. 
% The error is computed as the Euclidean norm (L2 norm) of the adjusted signals.
It is not necessary to perform the inverse GFT on the errors for the MSE calculation, as the norm is preserved under the Fourier transform. Consequently, the gradients computed in the frequency domain are equivalent to those computed in the time domain.:
\[
\hat{y}^{\text{train}'} = \text{adjust}(\hat{y}^{\text{train}}, \alpha), \quad \hat{y}' = \text{adjust}(\hat{y}, \alpha)
\]

\[
\text{MSE}_{\text{freq}} = \frac{1}{N} \|\hat{y}' - \hat{y}^{\text{train}'}\|_2^2
\]
where the \(\text{adjust}(\cdot)\) function operates on signals transformed to the frequency domain and multiplies their high-energy components with an adjustment factor \(\alpha\), which aims to balance the energy distribution across the spectrum. Further explanation can be found in Appendix \ref{sec:appendix_lambda}.

\section{Experiments}
\begin{figure}[htbp]
    \centering
    \includegraphics[width=0.85\linewidth]{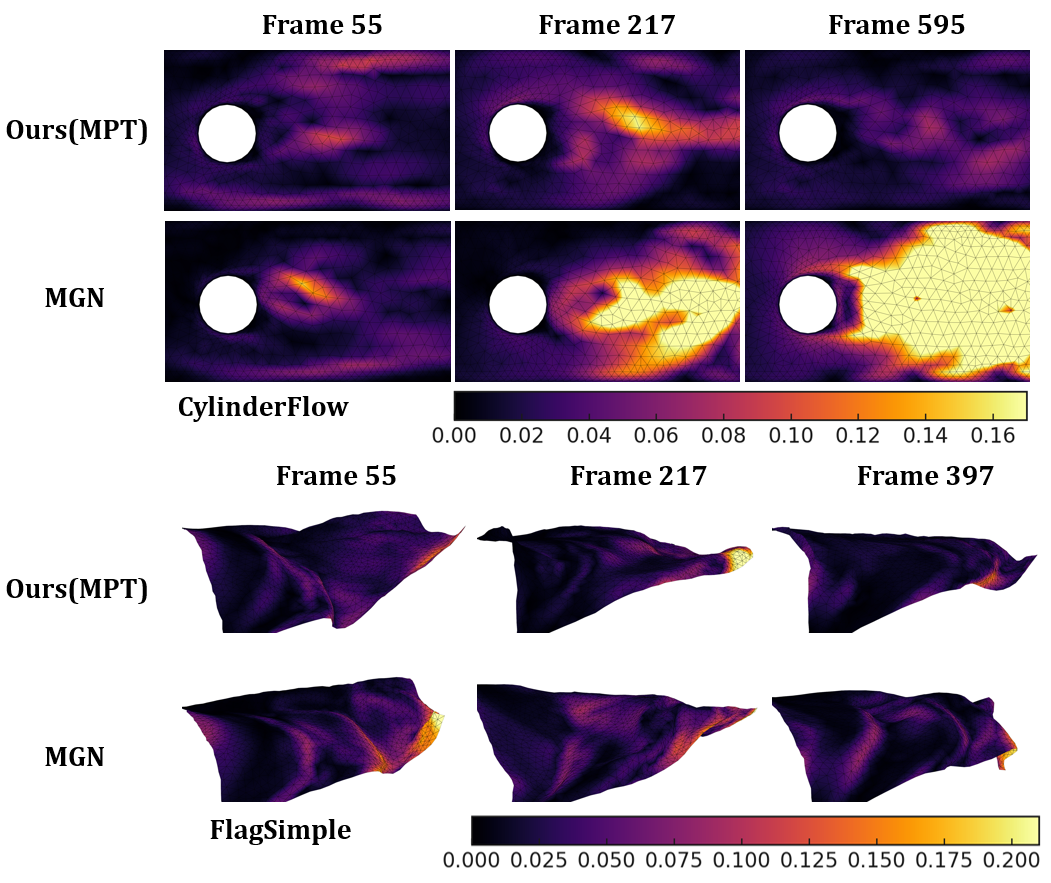}
    \caption{Comparison of RMSE of velocity norm between the Lagrangian system \textit{FlagSimple} and the Eulerian system \textit{CylinderFlow} using our MPT model and the MGN model \cite{MGN}.}
    \label{fig:compar}
\end{figure}
    In Section \ref{sec:task_setups}, we describe the datasets used and the implementation. We then describe baseline models for comparison and present and discuss the evaluation results and ablation studies conducted to assess the contributions of specific model components in Sections \ref{sec:eva} and \ref{par:ablation_studies}. We also visualize the RMSE of a Lagrangian system and an Eulerian system respectively, as shown in Figure \ref{fig:compar}.

    \subsection{Task setups}
    \label{sec:task_setups}
        \paragraph{Datasets Description}
            We evaluated our method in the representation of both Lagrangian and Eulerian dynamical systems. The Lagrangian systems involve the datasets \textit{FlagSimple} and \textit{DeformingPlate}, while the Eulerian systems include \textit{CylinderFlow} and \textit{Airfoil}, with all datasets sourced from MeshGraphNet \cite{MGN}.
            
            \begin{itemize}
                \item \textit{FlagSimple} models a flag blowing in the wind, utilizing  a static Lagrangian mesh with a static topology structure and ignores collisions.
                
                \item \textit{DeformingPlate} Utilizes a quasi-static simulator to model the deformation of a hyper-elastic plate by a kinematic actuator. The dataset is structured with a Lagrangian tetrahedral mesh.
                
                \item \textit{CylinderFlow} simulates the flow of an incompressible fluid around a fixed cylinder in a 2D Eulerian mesh.
                
                \item \textit{Airfoil} focuses on the aerodynamics around an airfoil wing section, employing a 2D Eulerian mesh to monitor the evolution of momentum and density.
            \end{itemize}

        \paragraph{Implementation}
            Our framework is built using PyTorch \cite{NEURIPS2019_bdbca288} and PyG (PyTorch Geometric) \cite{fey2019fast}. The entire model is trained and inferred on a single Nvidia RTX 4090. Detailed information, including network hyperparameters, input and output formats, and noise injection methods, can be found in appendix \ref{sec:MD}.

    \renewcommand{\arraystretch}{1.2}
    \begin{table}[htbp]
        \centering
        \begin{tabular}{|l|l l l l l|}
          \hline
          \textbf{Measurements} & \textbf{Dataset} & \textbf{\shortstack{Ours (MPT)}} & \textbf{\shortstack{MGN}} \cite{MGN} & \textbf{\shortstack{BSMS}} \cite{BSMS}& \textbf{\shortstack{TIE}} \cite{TIE}\\
          \hline
          RMSE-1 [1e-2] & Cylinder & \textbf{2.01E-01} & {5.83E-01} & {5.26E-01} & {4.97E-01} \\
                        & Airfoil & \textbf{{2.47E+02}} & {3.15E+02} & {3.05E+02} & {3.32E+02} \\
                        & Plate & \textbf{{1.03E-02}} & {2.67E-02} & {2.81E-02} & {3.54E-02} \\
                        & Flag & \textbf{{1.20E-02}} & {6.53E-02} & {6.98E-2} & {5.84E-2} \\
          \hline
          RMSE-50 [1e-2] & Cylinder & \textbf{6.33E-01} & {1.42} & {3.68} & {7.15} \\
                        & Airfoil & \textbf{{4.14E+02}} & {5.46E+02} & {1.37E+03} & {6.19E+03} \\
                        & Plate & \textbf{{9.26E-02}} & {1.73E-01} & {2.95E-01} & {3.58E-01} \\
                        & Flag & \textbf{{1.91}} & {2.28} & {2.67} & {2.33} \\
          \hline
          RMSE-all [1e-2] & Cylinder & \textbf{3.75} & {4.32} & {1.52E+1} & {2.89E+01} \\
                          & Airfoil & \textbf{{1.68E+03}} & {2.15E+03} & {1.03E+04} & {1.57E+05} \\
                          & Plate & \textbf{{1.13}} & {1.60} & {4.72} & {9.63} \\
                          & Flag & \textbf{{2.08}} & {2.57} & {3.58} & {1.24E+01} \\
          \hline
        \end{tabular}
        \caption{RMSE of our method, MeshGraphNet (MGN), Bi-Stride Multi-Scale GNN (BSMS-GNN) and Transformer with Implicit Edges (TIE) for different rollout steps. Our method achieves state-of-the-art in all datasets}
        \label{tab:eva}
    \end{table}
    \renewcommand{\arraystretch}{1.0}

    \subsection{Comparison with Baselines}
    \label{sec:eva}

        \paragraph{Baselines}
            In our evaluation, we compared against several state-of-the-art models. The Bi-Stride Multi-Scale Graph Neural Network (BSMS) \cite{BSMS} introduces multiscale methods to enhance the efficiency of message passing, significantly accelerating training and inference while maintaining simulation quality. MeshGraphNet (MGN) \cite{MGN} leverages a mesh-based approach for graph representation, enabling detailed and accurate physical simulations. Finally, the Transformer with Implicit Edges (TIE) model \cite{TIE} streamlines interaction modeling in Message Passing Neural Networks by utilizing a modified attention mechanism to efficiently process particle dynamics without explicit edge representations, improving computational efficiency. For the baseline models, we adopted the original parameter settings as specified in their respective papers.
            
        \paragraph{Evaluation}
            Table \ref{tab:eva} demonstrates the superiority of our model across all datasets. The \textit{CylinderFlow} dataset at RMSE-1 reveals a pronounced improvement with our model, which shows a reduction in error by \textbf{59.6\%} compared to TIE, the nearest competitor. For longer simulations, the comparative model shifts. At RMSE-50 and RMSE-all, MeshGraphNet (MGN) performs better than other baselines. Our model continues to exhibit superior performance, showing a reduction in error by \textbf{55.4\%} at RMSE-50 and by \textbf{13.2\%} at RMSE-all when compared to MGN. These improvements illustrate the robustness of our model, particularly in complex dynamic simulations where accurate initial predictions are of the utmost importance for the generation of reliable longer-term forecasts.
            
            In the context of the \textit{Airfoil} dataset, our model remains state-of-the-art. At the RMSE-50 condition, the model's error rate is reduced by \textbf{24.1\%} compared to MGN. This illustrates the model's capacity to maintain accuracy over prolonged sequences, which is an essential feature for simulations requiring stability over extended temporal spans. At RMSE-all, the improvement reaches its zenith with an error reduction of \textbf{21.8\%}.
            
            For the \textit{DeformingPlate} and \textit{FlagSimple} datasets, our model displays similar trends. In the Plate dataset, the RMSE-1 shows an improvement of \textbf{61.4\%} over MGN, with continued dominance in longer simulations, indicated by a \textbf{46.4\%} error reduction at RMSE-50. The Flag dataset demonstrates a more modest but consistent improvement across all metrics. The most notable reduction is \textbf{81.6\%} at RMSE-1.
            
            It is notable that TIE has higher errors due to the fact that it does not utilise edge information from the dataset. Instead, it computes attention over all points within a set window distance, which introduces approximation errors.

    \subsection{Ablation Studies}
    \label{par:ablation_studies}
        To rigorously evaluate the influence of specific model components and configurations on overall performance, systematic ablation studies were undertaken. These included: (1) Evaluating the contributions of individual components by comparing the effects of Hadamard-Product Attention (HPA) and Graph Fourier Loss (GFL), (2) comparing Dot-Product Attention with Hadamard-Product Attention, (3) assessing the efficacy of learnable lambda parameters $\lambda$ versus manual setting of $\lambda$, and (4) investigating how varying segmentation rates \(s_r\) affect model performance.
        
    \renewcommand{\arraystretch}{1.2}
    \begin{table}[htbp]
        \centering
        \renewcommand{\arraystretch}{1.2}
        \begin{tabular}{|l c c c c|}
          \hline
          \textbf{Measurements} & Without \textbf{HPA} and \textbf{GFL} & \textbf{HPA} only & \textbf{GFL} only & \textbf{HPA} + \textbf{GFL} (ours) \\
          \hline
          RMSE-1 [1e-2]         & 5.83E-1 & 2.64E-1 & 2.27E-1 & \textbf{2.01E-1} \\
          RMSE-50 [1e-2]        & 1.42    & 9.10E-1 & 6.96E-1 & \textbf{6.33E-1} \\
          RMSE-all [1e-2]       & 4.32    & 3.94    & 3.89    & \textbf{3.75} \\
          \hline
        \end{tabular}
        \caption{Ablation study conducted on the CylinderFlow dataset to evaluate the contributions of individual components within our architecture. We test the effects of Hadamard-Product Attention (HPA), Graph Fourier Loss (GFL), and their combination. Results indicate that integrating both HPA and GFL leads to reductions in error.}
        \label{tab:table3}
    \end{table}
    \renewcommand{\arraystretch}{1.0}

\begin{figure}[htbp]
    \centering
    \begin{subfigure}{0.329\textwidth}
        \centering
        \includegraphics[width=\textwidth]{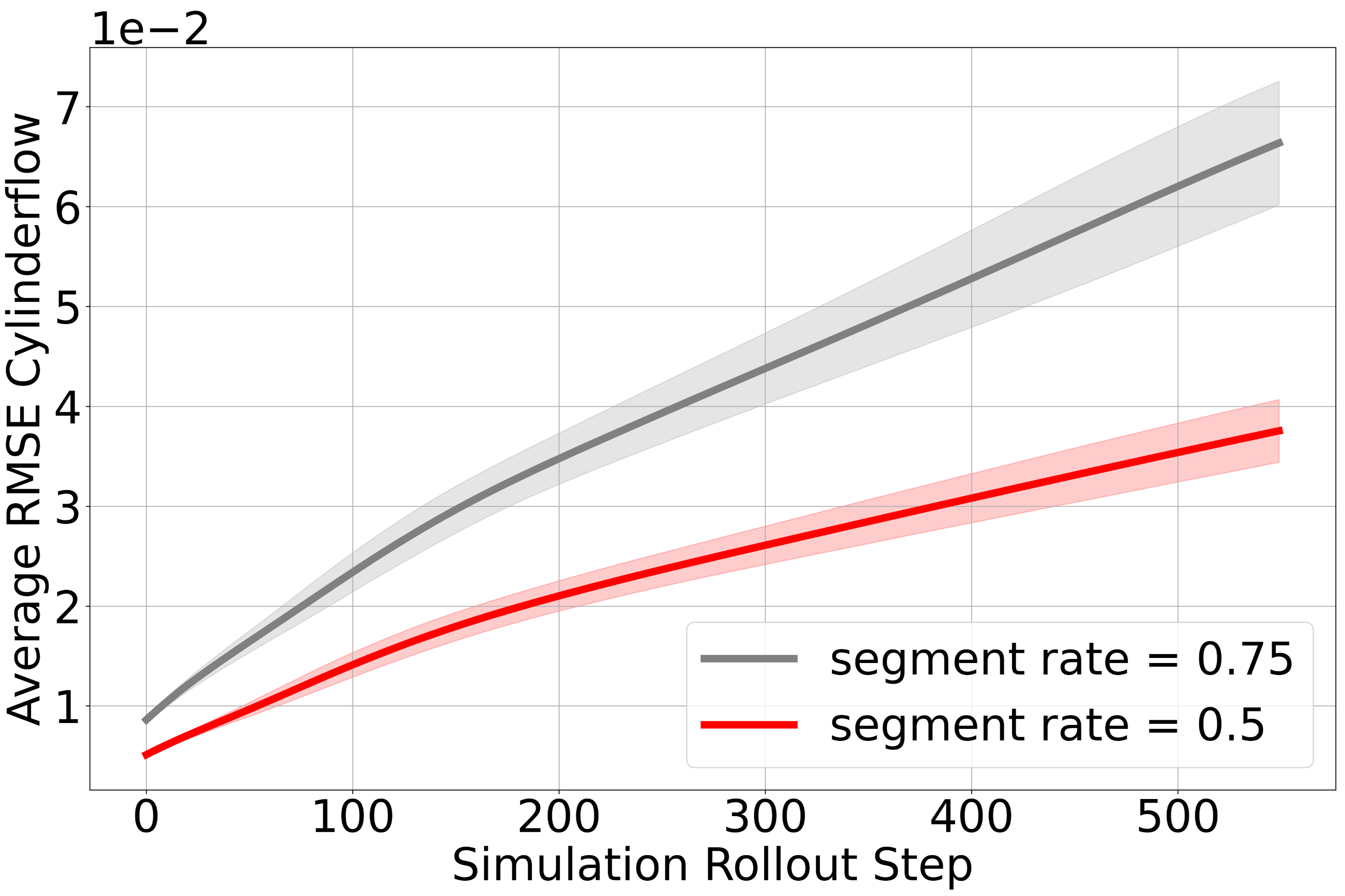}
        \phantomsubcaption
        \label{fig:DPA_vs_HPA}
        \text{(a)}
    \end{subfigure}
    \hfill
    \begin{subfigure}{0.329\textwidth}
        \centering
        \includegraphics[width=\textwidth]{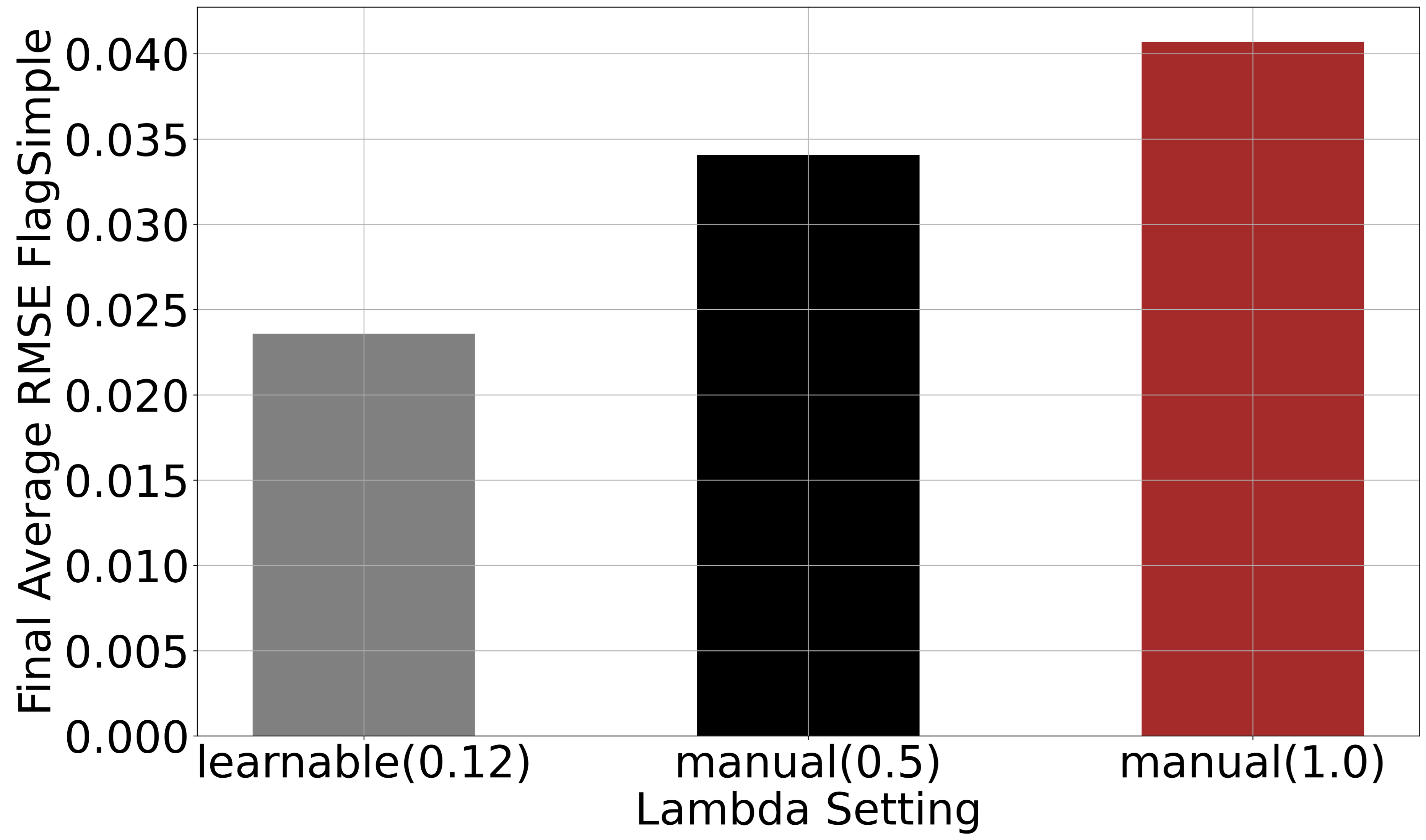}
        \phantomsubcaption
        \label{fig:lambda_comparison}
        \text{(b)}
    \end{subfigure}
    \hfill
    \begin{subfigure}{0.329\textwidth}
        \centering
        \includegraphics[width=\textwidth]{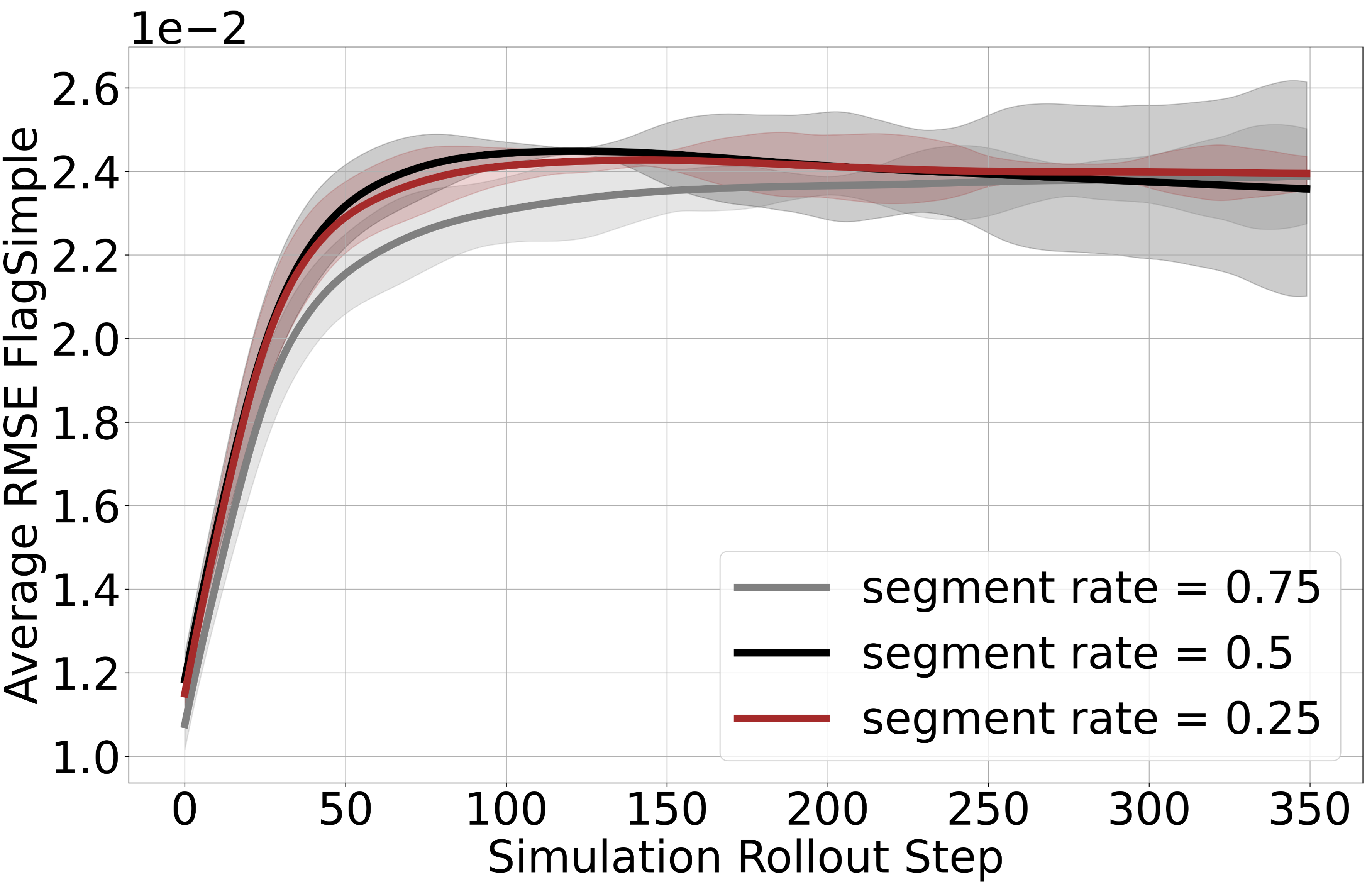}
        \phantomsubcaption
        \label{fig:seg_rate_impact}
        \text{(c)}
    \end{subfigure}
    \caption{(a) Comparison of Dot-Product Attention and Hadamard-Product Attention in \textit{CylinderFlow}. HPA demonstrates a lower average RMSE across all rollout steps compared to the Dot-Product Attention.
    (b) Comparison of learnable and manual $\lambda$ settings in \textit{FlagSimple}. Learnable $\lambda$ achieves lower error compared to manual settings.
    (c) The impact of varying segmentation rates \(s_r\) in \textit{FlagSimple}. Different segmentation rates do not significantly impact the final results.}
    \label{fig:comparisons}
\end{figure}

    \paragraph{Effectiveness of Graph Fourier Loss and Hadamard-Product Attention}
    In the ablation study focusing on the \textit{CylinderFlow} dataset, wwe test the effects of Hadamard-Product Attention (HPA), Graph Fourier Loss (GFL), and their combination on multiple dimensions of performance, including predictive accuracy as measured by RMSE.
    
     In terms of predictive accuracy, the model without HPA and GFL performed the worst, demonstrating significantly higher error rates across all RMSE measures. The integration of both HPA and GFL demonstrated the highest improvement in reducing error rates across all RMSE measures when compared to standalone implementations of GFL and HPA.
    
    \paragraph{Effectiveness of Hadamard-Product Attention}
    Figure \ref{fig:DPA_vs_HPA} compares the performance of Dot-Product Attention \cite{Transformer} and Hadamard-Product Attention (HPA) on the \textit{CylinderFlow} dataset. HPA consistently outperforms Dot-Product Attention across all rollout steps, demonstrating a lower average RMSE. This indicates that HPA's finer-grained feature dimension weighting is more effective in capturing the dynamics of the system, leading to more accurate predictions.
    
    \paragraph{Effectiveness of Learnable $\lambda$}
    Figure \ref{fig:lambda_comparison} compares the performance of models with learnable $\lambda$ settings against manual $\lambda$ settings in the \textit{FlagSimple} dataset. The results show that the learnable $\lambda$ achieves a lower final average RMSE compared to manual settings. This highlights the advantage of allowing the model to adaptively adjust $\lambda$ during training, leading to better overall performance.
    
    \paragraph{Segmentation Rate Selection}
    In Figure \ref{fig:seg_rate_impact}, we investigate the impact of varying segmentation rates \(s_r\) on the \textit{FlagSimple} dataset. The results indicate that different segmentation rates do not significantly impact the final results. This robustness to segmentation rate selection demonstrates that our model can maintain high performance regardless of the specific value of \(s_r\), simplifying the hyperparameter tuning process.
\section{Conclusion and Limitation}
The Message Passing Transformer (MPT) has achieved notable advancements in the accuracy of physical system simulations, leveraging Hadamard-Product Attention and Graph Fourier Loss to capture complex dynamics effectively. While MPT offers substantial improvements in simulation fidelity, it does face challenges with increased memory consumption, slower computational speeds, and limitations due to the fixed sequence length of the Hadamard-Product Attention. These issues, though significant, represent areas for future optimization and refinement. Continuing to develop MPT could lead to broader applications in dynamic system modeling and enhance its utility in scientific and engineering fields, driving forward the capabilities of learnable simulation technologies.

% \bibliographystyle{IEEEtran}
% \bibliography{citation}

\newpage
\section{Appendix}
\appendix
\section{Analysis of Learnable \(\lambda\) in Graph Fourier Loss}
    \label{sec:appendix_lambda}
    In our experiments, we observed a notable phenomenon: when using a learnable \(\lambda\) within the Graph Fourier Loss (GFL) framework, \(\lambda\) did not tend to zero. However, when applying a similar adjustment directly to the mean squared error (MSE) in the frequency domain, the value of \(\lambda\) quickly diminished to zero. This appendix section provides a comprehensive analysis of this observation and elucidates the underlying reasons.

    The key difference between the GFL approach and direct MSE adjustment lies in how \(\lambda\) interacts with the frequency domain energy components. In GFL, \(\lambda\) is indirectly involved through the calculation of an adjustment factor \(\alpha\), which is applied separately to the model's output \(y^{\text{train}}\) and the target output \(y\). This can be expressed as:
    \[
    \text{MSE}_{\text{freq}} = \frac{1}{N} \left\| \text{adjust}(U^T y^{\text{train}}, \alpha) - \text{adjust}(U^T y, \alpha) \right\|_2^2 
    \]
    In this formulation, \(\lambda\) helps balance the energy distribution across different frequency components, ensuring that it remains non-zero to maintain the desired balance between high and low-frequency components. The frequency domain energy for each signal is computed, and high and low-frequency components are adjusted based on \(\alpha\). Since \(\alpha\) is based on the mean energies and \(\lambda\) affects the balance of these means, the gradient of \(\lambda\) directly impacts the adjusted signals. This intricate adjustment mechanism precludes the possibility of a simple decline in the value of \(\lambda\), as it plays a pivotal role in maintaining equilibrium between the high and low-frequency energies.
    
    Conversely, when \(\lambda\) is directly applied to the MSE in the frequency domain using the combined error \(y^{\text{train}} - y\), the lack of separate intermediate adjustments for \(y^{\text{train}}\) and \(y\) leads to a different effect. This can be expressed as:
    \[
    \text{MSE}_{\text{freq}} = \frac{1}{N} \left\| \text{adjust}(U^T (y^{\text{train}} - y), \alpha) \right\|_2^2 
    \]
    Here, \(\lambda\) scales the error term without the intermediate balancing mechanism, causing \(\lambda\) to quickly diminish as it directly attempts to minimize the combined error in the frequency domain. When directly adjusting the error and computing the Fourier transform, \(\lambda\) primarily affects the adjustment of the error. As the model's prediction \(y^{\text{train}}\) and the target \(y\) become closer, the error \(y^{\text{train}} - y\) becomes smaller. In this scenario, the optimizer tends to reduce \(\lambda\) to zero to minimize the loss function.

\section{Dataset Details}
    \renewcommand{\arraystretch}{1.2}
    \begin{figure}[htbp]
    \centering
    \begin{minipage}[b]{0.45\textwidth}
        \centering
        \includegraphics[width=\textwidth]{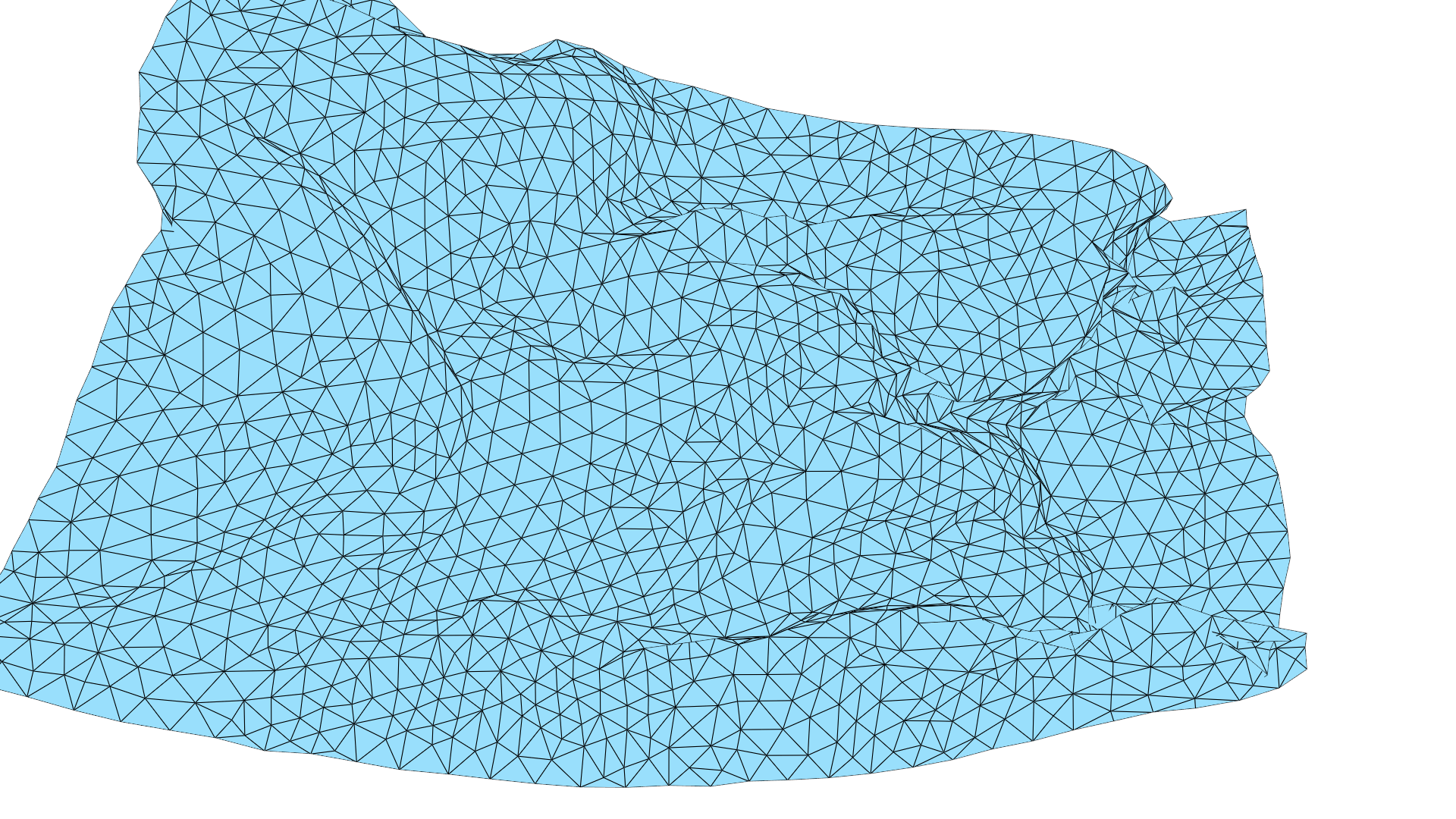}
        \subcaption{Flag Simple}
    \end{minipage}
    \begin{minipage}[b]{0.45\textwidth}
        \centering
        \includegraphics[width=\textwidth]{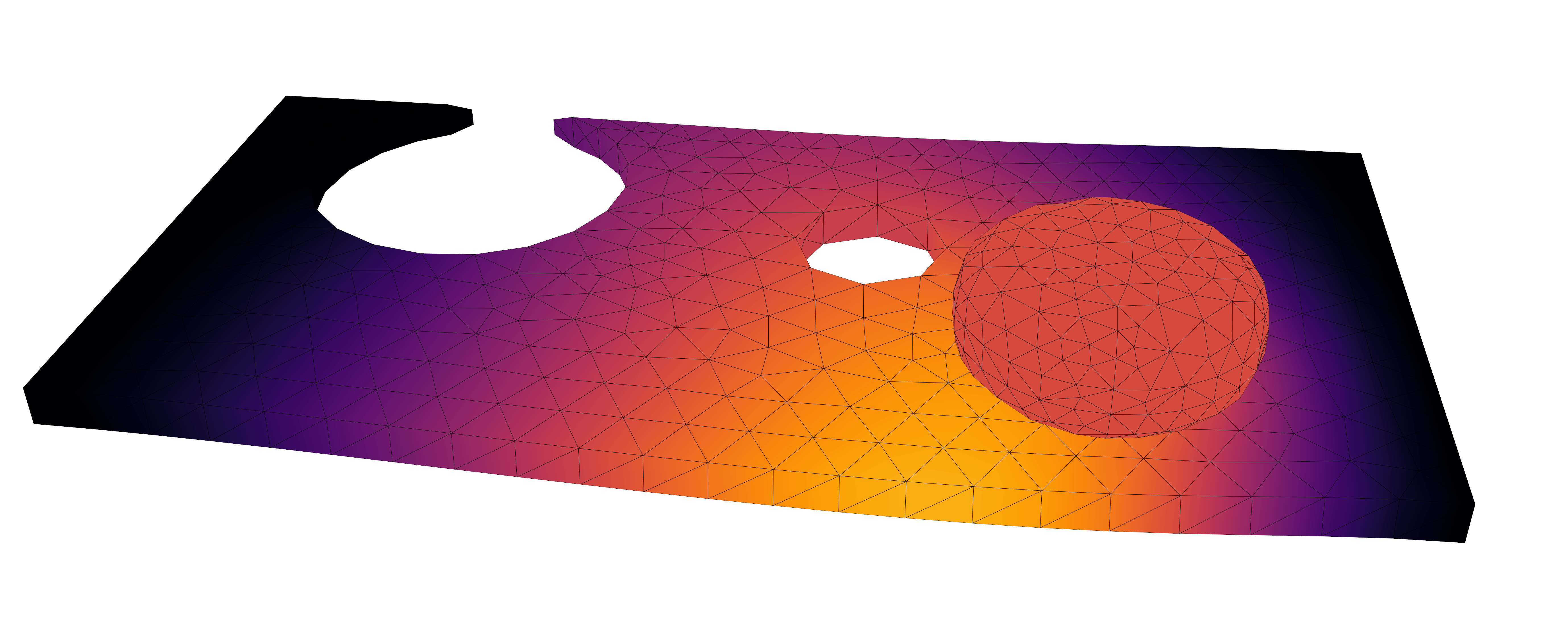}
        \subcaption{Deforming Plate}
    \end{minipage}
    \\
    \begin{minipage}[b]{0.45\textwidth}
        \centering
        \includegraphics[width=\textwidth]{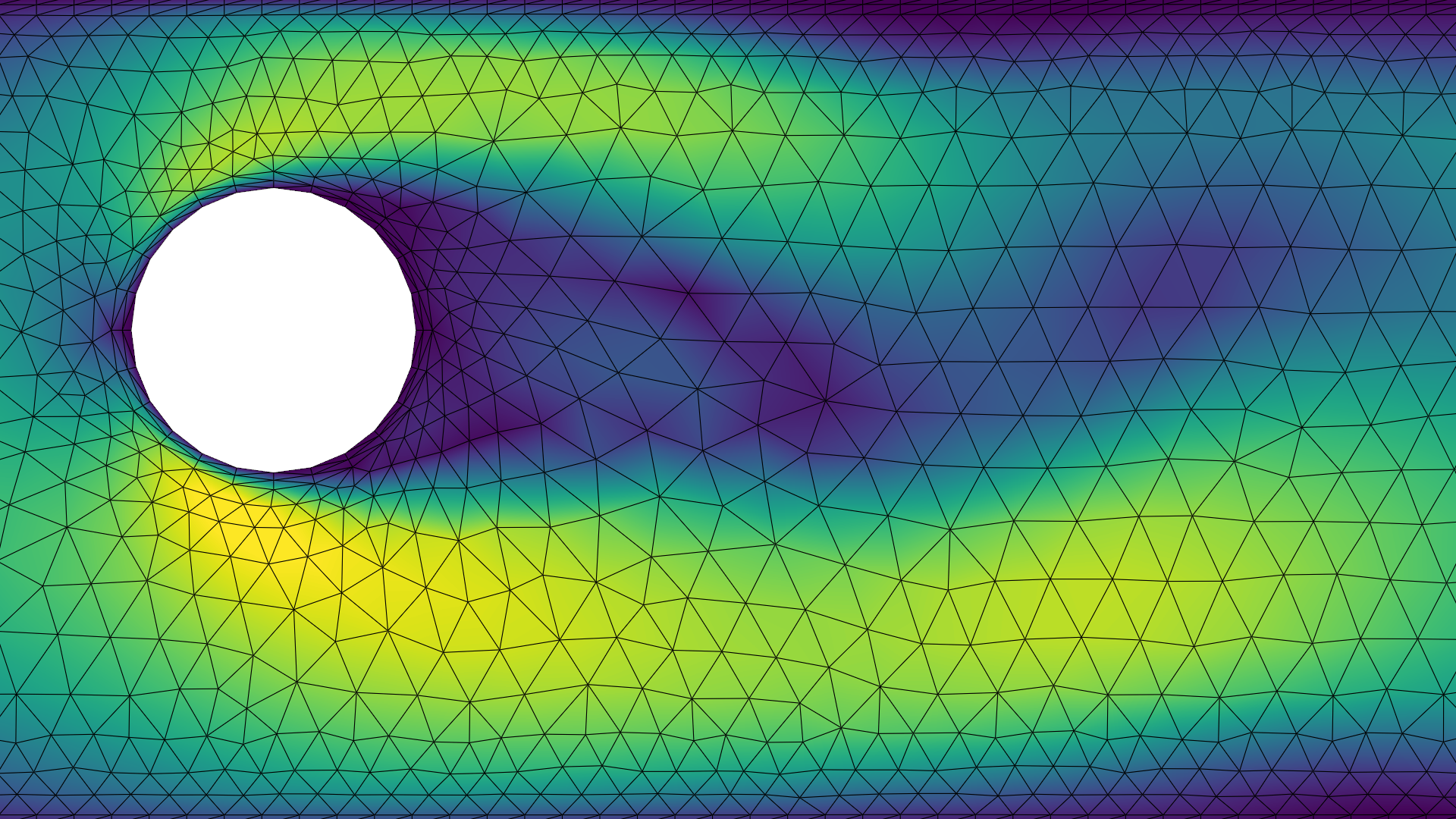}
        \subcaption{Cylinder Flow}
    \end{minipage}
    \begin{minipage}[b]{0.45\textwidth}
        \centering
        \includegraphics[width=\textwidth]{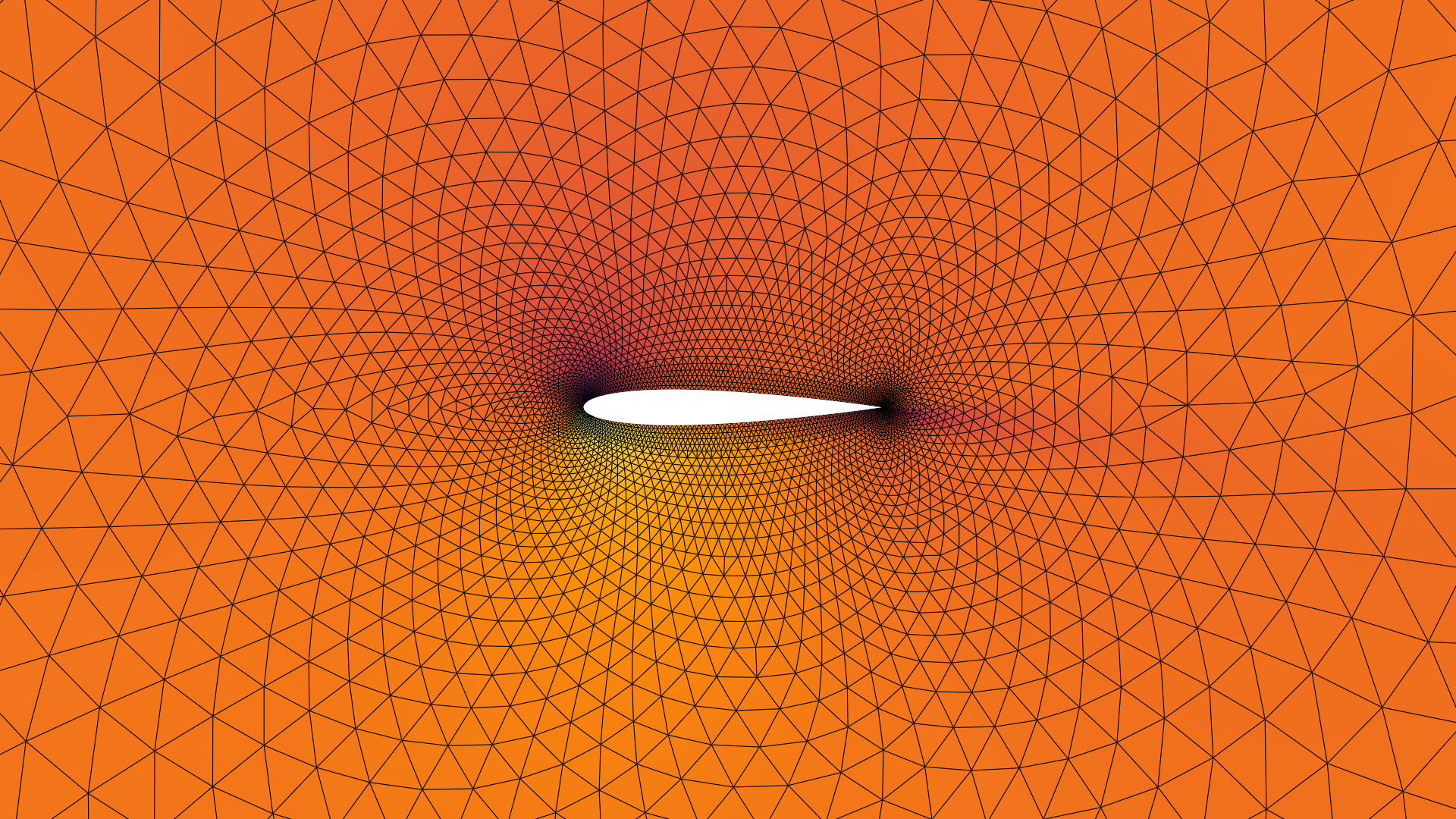}
        \subcaption{Airfoil}
    \end{minipage}
    
    \caption{Visualization of different datasets.}
    \label{fig:diff_datasets}
    \end{figure}
    
    \begin{table}[H]
    \centering
    \caption{Dataset Specifications}
    \label{tab:datasets}
    \begin{tabular}{|l|l|l|l|l|l|}
    \hline
    \textbf{Dataset} & \textbf{System} & \textbf{Mesh Type} & \textbf{Dimensions} & \textbf{\# Steps} & \textbf{time step $\Delta t$} \\ \hline
    FlagSimple       & Lagrangian      & triangle          & 3D                  & 400               & 0.02 \\ 
    DeformingPlate   & Lagrangian      & tetrahedral       & 3D                  & 400               & —    \\ 
    CylinderFlow     & Eulerian        & triangle          & 2D                  & 600               & 0.01 \\ 
    Airfoil          & Eulerian        & triangle          & 2D                  & 600               & 0.008 \\ \hline
    \end{tabular}
    \end{table}
    \renewcommand{\arraystretch}{1.0}
    
    \paragraph{Dataset Specifications}
    Our models are trained and evaluated across four distinct datasets: \textit{FlagSimple}, \textit{DeformingPlate},\textit{ CylinderFlow}, and \textit{Airfoil}. Each dataset consists of 1000 training trajectories, 100 validation trajectories, and 100 test trajectories, with each trajectory comprising between 250 to 600 time steps. The \textit{FlagSimple} dataset models a flag fluttering in the wind using a static Lagrangian mesh with a fixed topology, ignoring collision effects. The \textit{DeformingPlate} dataset simulates the deformation of a hyper-elastic plate driven by a kinematic actuator with a quasi-static simulator, structured on a Lagrangian tetrahedral mesh. The \textit{CylinderFlow} dataset involves the simulation of incompressible The \textit{Airfoil} dataset focuses on the aerodynamic properties around an airfoil section, utilizing a 2D Eulerian mesh to track changes in momentum and density over time. Table \ref{tab:datasets} details for each dataset include: \textbf{System}, indicating whether the simulation is Lagrangian for solid mechanics or Eulerian for fluid dynamics; \textbf{Mesh Type}, specifying the geometric configuration such as triangular or tetrahedral; \textbf{Dimensions}, indicating whether the simulation is in 2D or 3D; and \textbf{\# Steps}, the total number of simulation steps in each trajectory reflecting the depth of time-dependent analysis. The \textbf{time step $\Delta t$} column specifies the simulation time increment between each step.

\section{Model Details}
    \label{sec:MD}
    \paragraph{Model Hyperparameters}
        The training of our model employs a batch size of 1. The Message Passing (MP) time is all set at $15$ steps. For optimization, the Adam optimizer is used with an initial learning rate of $10^{-4}$, which decays exponentially to $10^{-6}$ over the course of 2 million training steps from a total of 5 million steps. The model employs four functions, namely, \(f_1\), \(f_2\), \(f_3\), and \(f_5\), which are all configured as ReLU-activated two-hidden-layer MLPs. Each of these MLPs has both a hidden and an output layer with a size of 128. The Hadamard-Product Attention mechanism implemented uses four heads and includes a dropout rate of $0.1$ to prevent overfitting. The segmentation rate, denoted by \(s_r\), is set at $0.5$. Rather than employing a manual setting of the parameter $\lambda$, we have chosen to utilize learnable lambda parameters, denoted by $\lambda$.
        
    \renewcommand{\arraystretch}{1.2}
    \begin{table}[H]
    \centering
    \caption{Model Input and Output Specifications}
    \label{tab:dataset_inputs_outputs}
    \begin{tabular}{|c|c|c|c|c|}
    \hline
    \textbf{Dataset} & \textbf{edge inputs $e^M_{ij}$} & \textbf{edge inputs $e^W_{ij}$} & \textbf{node inputs $v_i$} & \textbf{output} \\ 
    \hline
    FlagSimple       & $x_{m,ij}$, $|x_{m,ij}|$, \newline $x_{w,ij}$, $|x_{w,ij}|$       & $x_{w,ij}$, $|x_{w,ij}|$      & $n_i, \dot{x_i}$      & $\ddot{x}_i$ \\ 
    DeformingPlate  & $x_{m,ij}$, $|x_{m,ij}|$, \newline $x_{w,ij}$, $|x_{w,ij}|$       & $x_{w,ij}$, $|x_{w,ij}|$      & $n_i$                 & $\dot{x_i}, \sigma_i$ \\ 
    CylinderFlow    & $x_{w,ij}$, $|x_{w,ij}|$                                          & --                            & $n_i, w_i$            & $\dot{w_i}$ \\ 
    Airfoil         & $x_{w,ij}$, $|x_{w,ij}|$                                          & --                            & $n_i, w_i, \rho_i$    & $\dot{w_i}, \dot{\sigma_i}$ \\ 
    \hline
    \end{tabular}
    \end{table}
    \renewcommand{\arraystretch}{1.0}
    \paragraph{Model Input and Output}
    In table \ref{tab:dataset_inputs_outputs}, several specific terms and symbols define the structure of input and output data for each dataset involved in the simulations. The edge inputs \( e^M_{ij} \) and \( e^W_{ij} \) represent interactions associated with edges between nodes \(i\) and \(j\), where \(x_{m,ij}\) denotes the world edge position and \(x_{w,ij}\) indicates the mesh edge position. The node inputs \( v_i \) include \( n_i \), representing node types, and \( x_i \), indicating node positions. Other node-specific properties include momentum (\( w_i \)) and density (\( \rho_i \)), while outputs encompass acceleration (\( \ddot{x}_i \)), velocity (\( \dot{x}_i \)), and von Mises stress (\( \sigma_i \)).
    
    \renewcommand{\arraystretch}{1.2}
    \begin{table}[H]
    \centering
    \caption{Noise Scale and World Edge Radius Specifications}
    \label{tab:dataset_noise_radius}
    \begin{tabular}{|c|c|c|}
    \hline
    \textbf{Dataset} & \textbf{Noise Scale} & \textbf{World Edge Radius $r_w$} \\
    \hline
    FlagSimple & pos: 1e-3 & — \\
    DeformingPlate & pos: 3e-3 & 0.03 \\
    CylinderFlow & momentum: 2e-2 & — \\
    Airfoil & momentum: 1e1, density: 1e-2 & — \\
    \hline
    \end{tabular}
    \end{table}
    \renewcommand{\arraystretch}{1.0}
    \paragraph{Noise Injection and World Edge Radius Settings}
    To enhance the robustness of our model against noisy inputs and to simulate real-world data conditions more accurately, we implemented a strategy for noise injection into the training process. These noise scales are consistent with the settings used in MeshGraphNet \cite{MGN}, as detailed in Table \ref{tab:dataset_noise_radius}. Additionally, the \textbf{world edge radius $r_w$} column specifies the radius used for defining mesh interactions in the DeformingPlate dataset.

\section{Rollout Visualizations}
\begin{figure}[htbp]
    \centering
    \includegraphics[width=\linewidth]{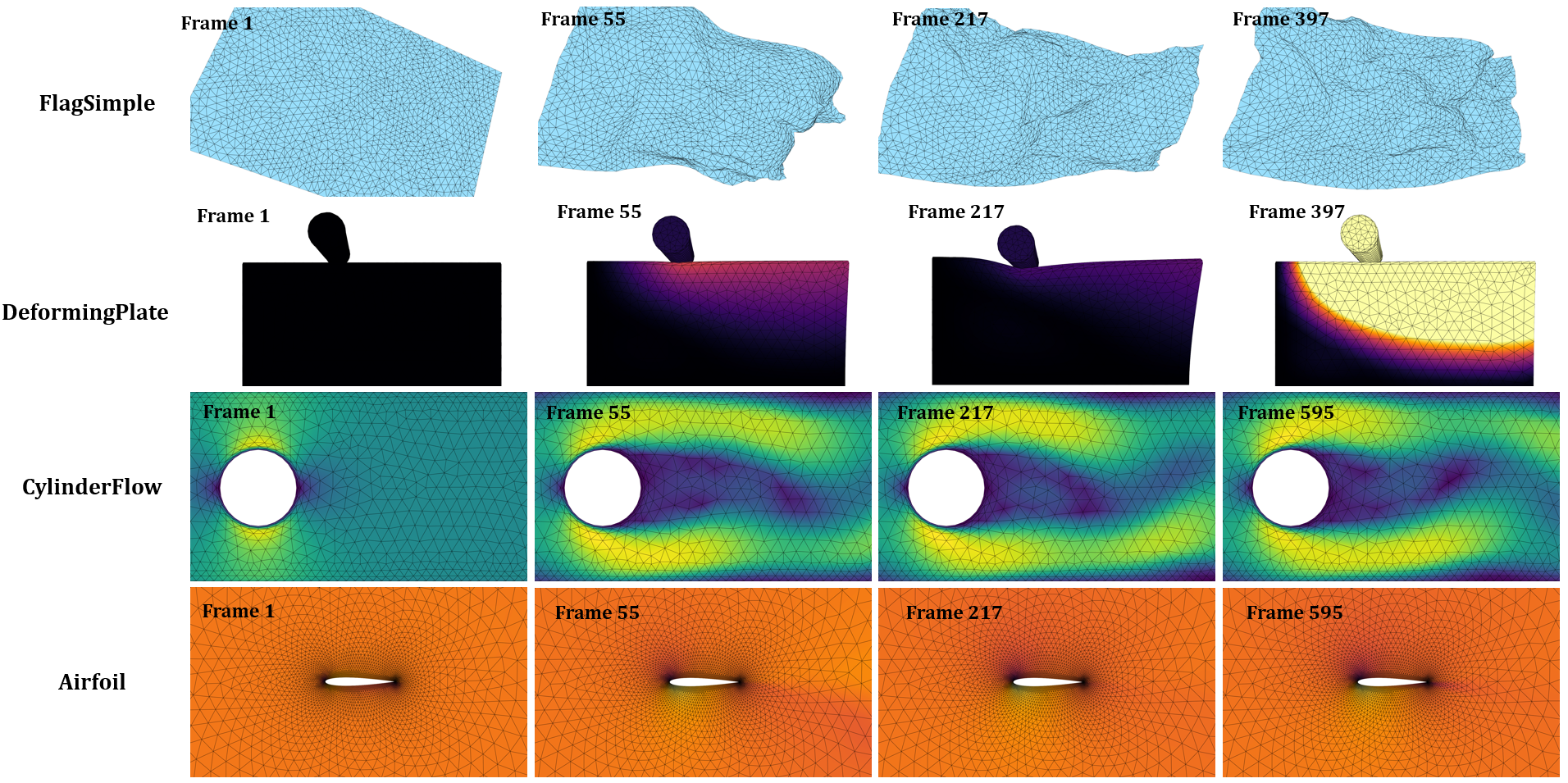}
    \caption{Rollout results of MPT: the \textit{FlagSimple} visualizes only the world position, while other colors represent the velocity norm.}
    \label{fig:compar}
\end{figure}

\end{document}